\documentclass[final]{cvpr}

\usepackage{times}
\usepackage{epsfig}
\usepackage{graphicx}
\usepackage{amsmath}
\usepackage{amssymb}
\usepackage{subfig}
\usepackage{mathrsfs}
\usepackage{blindtext}
\usepackage[english]{babel}
\usepackage{color}
\usepackage[noend]{algpseudocode}
\usepackage{algorithmicx,algorithm}
\usepackage[shortlabels]{enumitem}
\usepackage{multirow}
\usepackage{amssymb}
\usepackage[titletoc]{appendix}

\usepackage[pagebackref=true,breaklinks=true,colorlinks,bookmarks=false]{hyperref}

\def\cvprPaperID{2397} 
\def\confYear{CVPR 2021}

\begin{document}

\title{Online Learning of a Probabilistic and Adaptive Scene Representation}

\author{Zike Yan  \qquad \qquad \qquad Xin Wang \qquad \qquad \qquad Hongbin Zha\\
	Key Laboratory of Machine Perception (MOE), School of EECS, Peking University\\
	PKU-SenseTime Machine Vision Joint Lab\\
	{\tt\small zike.yan@pku.edu.cn  \qquad xinwang\_cis@pku.edu.cn  \qquad zha@cis.pku.edu.cn}
}

\twocolumn[{%
	\renewcommand\twocolumn[1][]{#1}%
	\maketitle
	\begin{center}
		\centering
		\begin{minipage}{.9\linewidth}
			\centering
			\includegraphics[width=.8\textwidth]{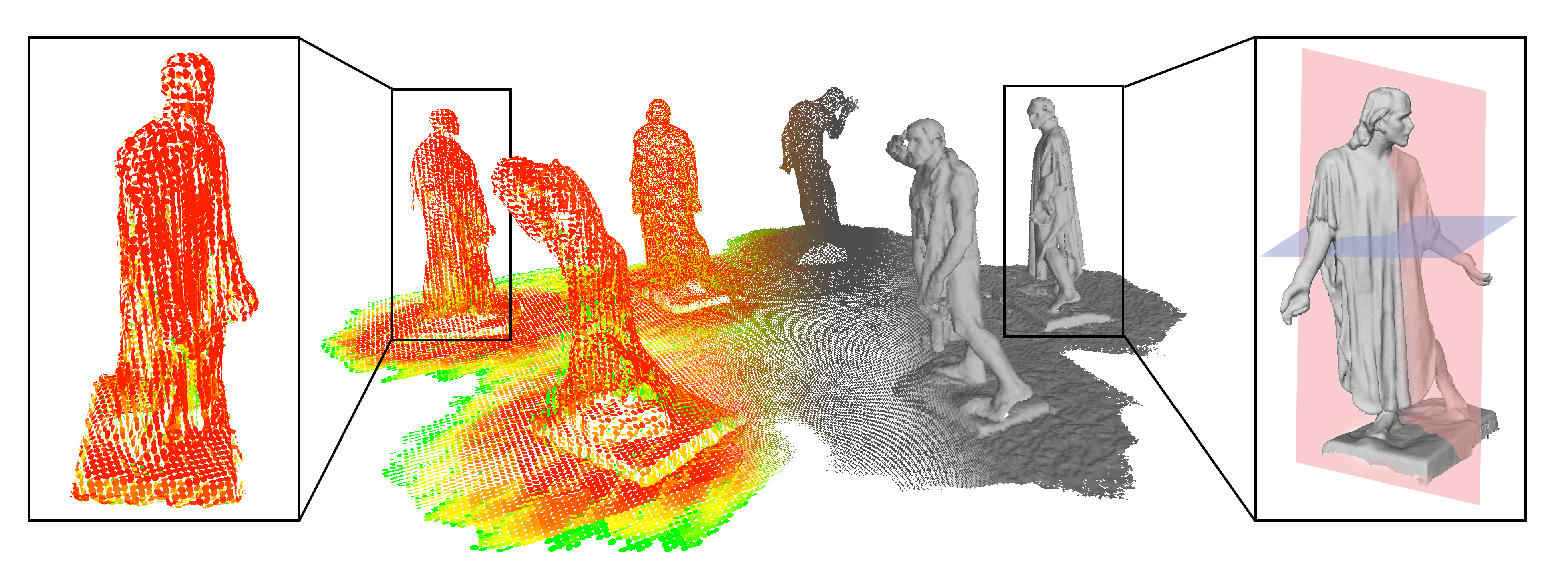}
			\label{fig:title}
		\end{minipage}
		\captionof{figure}{We propose a continuous probability field that can be learned incrementally from streaming data. The probabilistic formulation naturally incorporates both geometry and uncertainty information into a compact parameter space. The generative characteristic allows convenient conversion to different kinds of scene representations.}
	\end{center}%
}]


\newcommand{\setmode}[1]{\def\mode{#1}}
\setmode{draft} 
\long\def\IGNORE#1{} \long\def\COMMENT#1{}
\def\authornote#1#2#3{{\textcolor{#2}{\textsl{\small#1:[*#3*]}}}}
\ifthenelse{\equal{\mode}{draft}} {
	\newcommand{\zknote}[1]{\authornote{ZK}{red}{#1}} 
} {}

\ifthenelse{\equal{\mode}{final}} {
	\newcommand{\zknote}[1]{}
	\typeout{************* MODE: Final}
} {}

\begin{abstract}
	Constructing and maintaining a consistent scene model on-the-fly is the core task for online spatial perception, interpretation, and action. In this paper, we represent the scene with a Bayesian nonparametric mixture model, seamlessly describing per-point occupancy status with a continuous probability density function. Instead of following the conventional data fusion paradigm, we address the problem of online learning the process how sequential point cloud data are generated from the scene geometry. An incremental and parallel inference is performed to update the parameter space in real-time. We experimentally show that the proposed representation achieves state-of-the-art accuracy with promising efficiency. The consistent probabilistic formulation assures a generative model that is adaptive to different sensor characteristics, and the model complexity can be dynamically adjusted on-the-fly according to different data scales.
\end{abstract}


\section{Introduction}
\label{sec:intro}
Simultaneous Localization and Mapping (SLAM) has been recently viewed as a potential perceptual tool towards \emph{Spatial AI}~\cite{Davison2018arXiv} as it allows a mobile device to perceive the world and estimate the sensor state. Along with the evolution of SLAM systems towards spatial perception arises an increasing demand for a more expressive map that can incrementally distill knowledge from different kinds of data into a compact parameter space. Finding an appropriate representation has been a central task of establishing such a comprehensive map. 

In this paper, we aim to maintain a continuous probability field that allows for storing data into a unified probabilistic form. The probability field offers a generative extension to different spatial representations, \eg., point cloud, occupancy grid, mesh at arbitrary resolution. Practically, we propose a continuous probability density function as the map representation using a Bayesian nonparametric mixture model. When obtaining 3D point cloud data, the scene geometry is depicted as a continuous probability field of spatial occupancy status. This representation owns the following properties: \emph{1) Probabilistic.} The Bayesian fashion not only quantifies uncertainties explicitly, but also allows to incorporate all sorts of information from different sensor inputs in a unified probabilistic manner; \emph{2) Adaptive and dynamic.} The nonparametric property offers an inherently infinite capacity~\cite{Xuan2019csur} that guarantees an adaptive model complexity with respect to acquired data scale; \emph{3) Compact and expressive.} The mixture model maintains a continuous and dense probability field in a discrete and sparse parameter space, where both geometry and uncertainty are kept within Gaussian components.

Specifically, we formulate the mapping as an online Bayesian learning problem: the map provides a generative process of the observations, and we use acquired streaming data to learn it incrementally. The incremental inference can be viewed as a transition from geometry prior to posterior given streaming data. As the posterior is intractable to compute and represent, we resort to a parallel and incremental approach. The global distribution is parallelly distributed to local processing in an incremental fashion, guaranteeing efficient inference for accurate scene geometry. 

In summary, our major contributions include a novel scene representation using the Bayesian nonparametric mixture model and a principled way of online Bayesian learning for efficient map updating. Our method obtains a continuous high-quality scene representation incrementally, and achieves state-of-the-art results as demonstrated in the qualitative and quantitative experiments.
\section{Related Work}
\label{sec:relatedwork}
In this paper, we introduce a novel scene representation that is probabilistic, adaptive, and can be learned incrementally from sequential data. Here we review the most closely related scene representations and indicate the major differences between ours.
\subsection{3D Scene Representations} 
Commonly-used 3D scene representations can be broadly categorized into three kinds: global function based, local primitive based, and neural representations.

\noindent\textbf{Global function based.} 
Global function based approaches represent the scene geometry as a continuous scalar field and maintain a global function to map \emph{xyz} coordinate to the field. Signed distance function (SDF) is one commonly used implicit function to represent zero level-set surfaces. Prevalent approaches~\cite{Curless1996siggraph, Kinectfusion} tend to discretize the space into regularly-partitioned voxel grids and directly maintain a discrete signed distance field. Though this volumetric representation is easy to manage and allows convenient rendering and data fusion, the signed distance function based approaches rely highly on the voxelization as it contains barely geometric property, hence struggling against scalability and flexibility~\cite{Chen2013tog, VoxelHashing, Whelan2015ijrr, Dai2017tog}. Besides, the continuity of the distance function is broken due to voxelization.

Another kind of approach represents the scene with a continuous probability density function (PDF) to maintain per-point occupancy probability, which is similar to our approach. Commonly used representations include Gaussian mixture model (GMM)~\cite{Srivastava2018tro, Eckart2016cvpr,Keselman2019crv,Meadhra2018ral, Tabib2019rss} and Gaussian process (GP)~\cite{Callaghan2010icra,Callaghan2012ijrr, Martens2016ral,Wang2016icra, Jadidi2017ral}. GMM is commonly used as a compact generative model for scene geometry. The uncertainty-aware nature makes it appropriate for robust point cloud registration~\cite{Eckart2018eccv, Gao2019cvpr, Evangelidis2017pami}. However, GMM requires a pre-defined number of mixtures, which is non-trivial to be applied for sequential data. On the other hand, Gaussian process is a Bayesian nonparametric model that is closely related to ours. Mapping with a Gaussian process is cast as a surface function regression problem. A similar idea is applied to Hilbert map~\cite{Ramos2016ijrr, Guizilini2018ijrr, Senanayake2017corl} that projects observations into a reproducing kernel Hilbert space. However, online operation with the Gaussian process representation is restrained by the requirement to cache training data during inference and the computationally-burdensome inversion of covariance matrix~\cite{Srivastava2018tro}. In contrast, we achieve real-time performance through incremental and parallel inference.

\noindent\textbf{Local primitive based.} 
Local primitive based approaches represent a scene with a set of discrete geometric primitives. Inference on this kind of representation is performed by fitting local geometry, usually planar surfaces, with the primitives. Commonly used primitives include surfel~\cite{Whelan2015rss, Schops2019cvpr, Keller2013_3dv}, mesh triangle~\cite{Delaunoy2014cvpr, Delaunoy2011ijcv}, voxel grid~\cite{Hornung2013ar}, and 3D Gaussian (ellipsoid)~\cite{Dhawale2020rss}.

Surfel~\cite{surfels} represents local geometry as an oriented disk. The unstructured nature makes it flexible for deformation and adaptive to different geometric frequencies. However, surfel is inherently sparse, thus leading to a discrete and incomplete scene model. Mesh provides a watertight surface model that is applicable for action and rendering. However, the topology changes for mesh representation are computationally expensive. Incremental mesh extraction is usually derived from other representations such as volumetric SDF~\cite{Dong18icra} and surfel~\cite{SurfelMeshing}.

Another line of research maintains local occupancy status within a sparsely partitioned area. Octomap~\cite{Hornung2013ar} represents local geometry with uncertainty-aware voxel grids. The uncertainty of occupied, free, or unknown status is assumed to be consistent within a voxel. \cite{Dhawale2020rss} further maintains a set of unstructured ellipsoids parametrized by 3D Gaussians. Local geometry is assumed to share the same spatial distribution, where each ellipsoid is a 3D probabilistic extension of the surfel primitive. Normal distributions transform (NDT)~\cite{Biber2003iros, Saarinen2013icra, Saarinen2013ijrr, Schulz2018iros} can be viewed as a combination of voxel grid and 3D Gaussian. The occupancy status within a voxel is no longer a single scalar value but a more expressive Gaussian distribution. Though NDT is usually defined as a continuous representation from a voxelized GMM perspective, the voxel-wise local processing lacks a global constraint.

\noindent\textbf{Neural representations.}
Neural representations learn to parameterize the shape manifold with neural networks. The insights behind neural representations usually derive from a view synthesis perspective~\cite{Mildenhall2020eccv, Niemeyer2020cvpr, Sitzmann2019nips} or from conventional representations mentioned above, \eg., DeepSDF~\cite{Park2019cvpr}, PointGMM~\cite{Hertz2020cvpr}, ONet~\cite{Mescheder2019cvpr}, LDIF~\cite{Genova2020cvpr}. The network is expected to learn class-specific shape priors that allow shape completion, interpolation, and generation. Though most works in the area are restricted to an object-level reconstruction, progress has been made recently that achieves detailed scene-level reconstruction~\cite{Jiang2020cvpr, Peng2020eccv, Chabra2020eccv}. However, these approaches are prevented from online operation with sequential data due to the batch-training fashion. Our method, on the other hand, adopts an efficient and incremental inference that resorts to an uncertainty-aware and interpretable Bayesian learning fashion~\cite{Xuan2019csur}.

\subsection{Probabilistic 3D Data Fusion in Real-time}
3D sequential data are usually redundant and noisy. To ensure scalable exploration and real-time action capability for geometry-dependent mobile devices, probabilistic fusion is performed to compress observed noisy data into a clean and compact form. Acquired data are usually assumed as Gaussian-distributed noisy observations. Hence, weighted averaging is required to incrementally update the representation parameters according to the data. Voxel-based representations, \eg., volumetric TSDF~\cite{Curless1996siggraph}, NDT~\cite{Biber2003iros}, occupancy grid~\cite{Hornung2013ar}, assign each point to a voxel to update the corresponding geometric property, while unstructured representations such as surfel~\cite{Whelan2015rss} and 3D Gaussian~\cite{Dhawale2020rss} assign each point to a geometric primitive through projective association.

Follow-ups further improve the robustness against noise and outliers by designing more reasonable weight calculations or introducing more complex distributions over the parameter space. Yan \etal~\cite{Yan2017dense} encode uncertainties into the surfel map by maintaining a 3D positional covariance and a 1D illuminational covariance. Lee \etal~\cite{Lee2019icra} utilize a more expressive Gaussian process over the SDF value to maintain a continuous implicit surface function. Dong \etal~\cite{Dong2018eccv} add additional uniform distribution to handle outliers and explicitly model directional sensor noise. The literature in sensor measurement model~\cite{Nguyen2012Kinect, Huber2018icra, Khoshelham2012accuracy, Mallick2014sensor} is also vast, but the field is beyond our scope. Recently, RoutedFusion~\cite{Weder2020cvpr} proposes a 2D depth routing network and a 3D depth fusion network to learn non-linear TSDF updates in real-time and achieves state-of-the-art performance. 

We, on the other hand, share a similar idea with~\cite{Woodford2012eccv} to learn a generative model of the observation process. By directly modeling a continuous spatial distribution, the uncertainty-aware characteristic is naturally incorporated in a theoretically-principled way, and a probabilistic framework is established systematically.
\section{Overview}
In this section, we introduce the general idea of how the proposed representation is learned incrementally in real-time. The mathematical formulation from an online learning perspective is first presented, followed by a scene representation definition. A parallel and incremental scheme for efficient inference is then introduced.
\subsection{Problem Formulation}
We aim to maintain a spatial distribution $\mathbf{G}$ to represent the scene geometry.
Let $\mathbf{X} = \{\mathbf{X}^1,\mathbf{X}^2, \cdots, \mathbf{X}^t, \cdots\}$ be the streaming observations, where each set $\mathbf{X}^t$ consists of $N^t$ data points $\mathbf{x}_i^t \in \mathbf{X}^t$. We assume that observed data are i.i.d. samples drawn from the global distribution. The objectiveness is then to maintain and update a parameter space $\boldsymbol{\theta}_k^t \in \mathbf{\Theta}^t$ incrementally as the measurement of the spatial distribution. $\mathbf{\Theta}^t$ can be estimated by computing the posterior through Bayesian theorem recursively as:
\begin{equation}
	\label{eq:posterior}
	\begin{aligned}
		p(\mathbf{\Theta}^t|\mathbf{X}^{1:t}) =\frac{p(\mathbf{X}^t|\mathbf{\Theta}^t)p(\mathbf{\Theta}^t|\mathbf{X}^{1:t-1})}{p(\mathbf{X}^t|\mathbf{X}^{1:t-1})}.
	\end{aligned}
\end{equation} 

Under a Markov assumption, $\mathbf{X}^t$ is independent of $\mathbf{X}^{1:t-1}$. The posterior can then be transformed as:

\begin{equation}
	\label{eq:onlinelearning}
	\begin{aligned}
		p(\mathbf{\Theta}^t|\mathbf{X}^{1:t}) \propto \prod_{i=1}^{N^t} p(\mathbf{x}^t_i|\mathbf{\Theta}^t) p(\mathbf{\Theta}^t).	
	\end{aligned}
\end{equation} 

Through Eq.~\ref{eq:onlinelearning}, online Bayesian learning can be understood as a gradual transition from the geometric prior $p(\mathbf{\Theta}^t)$ to the posterior $p(\mathbf{\Theta}^t|\mathbf{X}^{1:t})$. Knowledge is incrementally learned from data, describing the generative process of streaming observations $\mathbf{X}^{1:t}$ under the routine of Bayesian theorem.




\subsection{Scene Representation}
\begin{figure}[t]
	\begin{center}
		\includegraphics[width=0.8\linewidth]{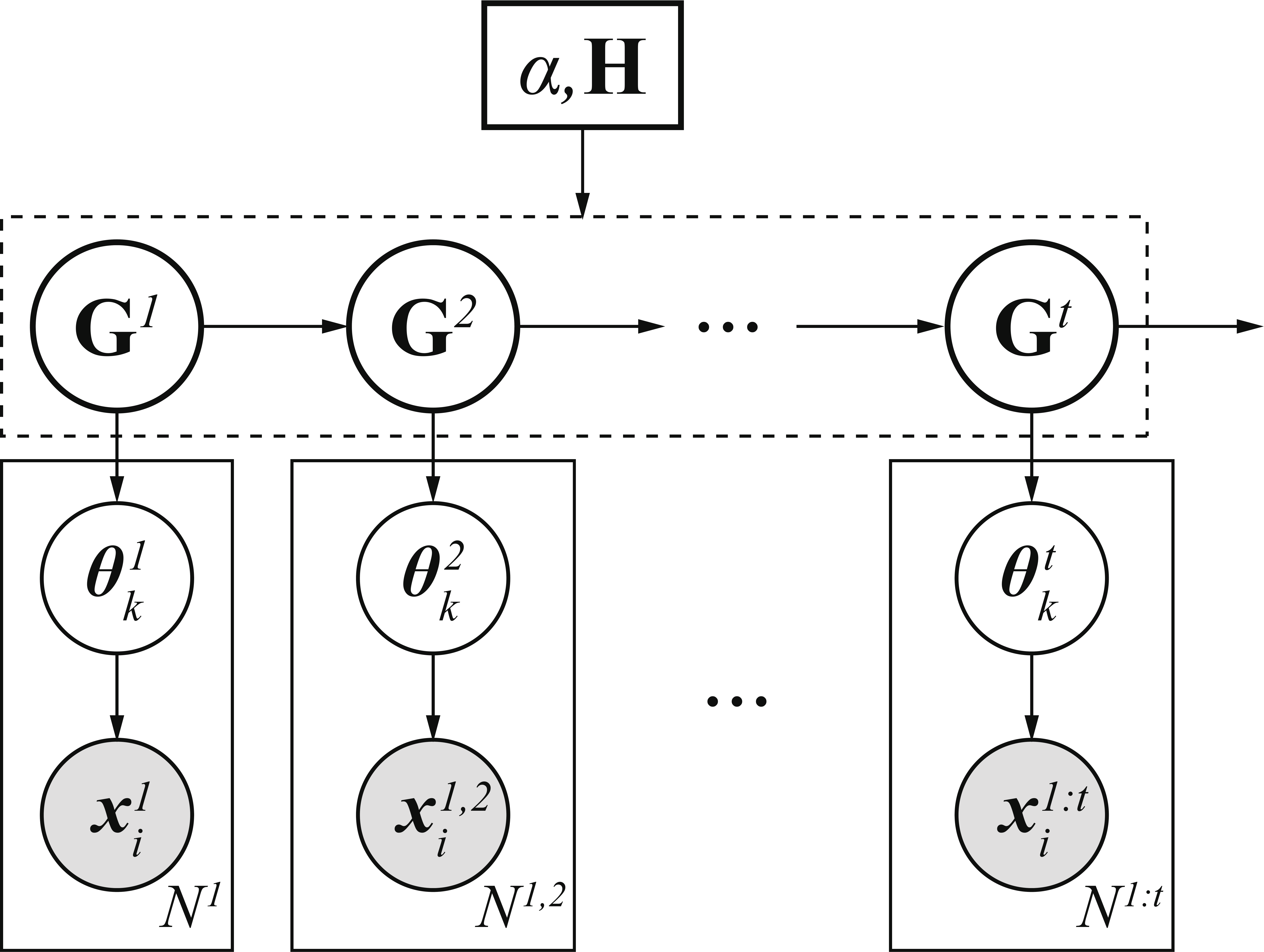}
	\end{center}
	\caption{The construction process of the proposed mixture model. Arrows indicate the conditional dependence. Hyper-parameters $\{\alpha, \mathbf{H}\}$ enforce a globally-consistent constraint. The measured probability field $\mathbf{G}^t$ progressively evolves as new data streamed in.}
	\label{fig:graph_model}
\end{figure}
In this paper, we introduce a Dirichlet process (DP) mixture model as the scene representation. As illustrated in Fig.~\ref{fig:graph_model}, the generative procedure of observations is well-explained by the model construction as:
\begin{equation}
\label{eq:construction}
	\mathbf{x}_i^t \sim \boldsymbol{\theta}_k^t, \boldsymbol{\theta}_k^t \sim \mathbf{G}^t,  \mathbf{G}^t \sim {\rm DP}(\alpha, \mathbf{H}),
\end{equation}
where $\boldsymbol{\theta}_k^t \in \mathbf{\Theta}^t$ is the $k$th mixture component parameterized by $[\omega_k^t,\boldsymbol{\mu}_k^t,\boldsymbol{\Sigma}_k^t]$. $\mathbf{G}^t$ is the global distribution measurement at time $t$. The concentration parameter $\alpha$ determines the sensitivity of component instantiation: the larger
$\alpha$ leads to an easier instantiation strategy. On the other hand, the base distribution $\mathbf{H}$
determines the initialization of the newly-instantiated component.

In the view of mixture construction, $\mathbf{G}^t$ is a distribution over sparse partitions~\cite{Pitman2002report} and can be discretized into countably-infinite components. Hence, it is usually viewed as an infinite-dimensional extension of the Gaussian mixture model ~\cite{Rasmussen2000nips}. The model can be intuitively understood using a Chinese restaurant process (CRP) metaphor: the $i$th customer $\mathbf{x}_i^t$ walks into a Chinese restaurant with an infinite number of tables and choose to sit at an already occupied table $\boldsymbol{\theta}_k^t$ or a new table $\boldsymbol{\theta}_{K^t+1}^t$. For our case, the sequential construction of the mixture model is handled with dynamic components creation and deletion~\cite{Lin2013nips}, thus leading to an adaptive model complexity according to the data scale.

\subsection{Online Bayesian Learning}
One bottleneck for Bayesian nonparametric learning lies in the fact that our objective posterior is intractable to compute and represent. We here resort to a parallel and incremental inference: the streaming data can be viewed as a sequence of mini-batches that arrive at consecutive epochs~\cite{Campbell2015nips, Ahmed2008sdm}. Each subset of data is then assigned to a thread-safe processing unit for local inference. Hence, the Dirichlet process mixture model is re-parameterized as a mixture of DPs, where inference on each DP is performed in parallel with the associated mini-batch data stream.

Let $\pi_i^t=j$ be the processor indicator for each observation $\mathbf{x}_i^t$ and $J^t$ be the number of processors to be allocated at time $t$. Assuming that data inside each epoch are exchangeable~\cite{Ahmed2008sdm} and thus conditionally independent, our objective posterior in Eq.~\ref{eq:posterior} can be decomposed into multiple local DPs. Following AVparallel~\cite{Williamson2013icml}, the generative procedure of the mixture model in Eq.~\ref{eq:construction} can be re-written as a mixture of DPs: 
\begin{equation}
\mathbf{G}^t=\sum\phi_j\mathbf{G}_j^t\sim{\rm DP}(\sum{\alpha_j}, \frac{\sum\alpha_j\mathbf{H}_j)}{\sum\alpha_j},
\end{equation}
where the construction of each DP is formulated as:
\begin{equation}
	\label{eq:parallel}
	\mathbf{x}_i^t \sim \boldsymbol{\theta}_{jk}^t, \boldsymbol{\theta}_{jk}^t \sim \mathbf{G}_j^t,  \mathbf{G}_j^t \sim {\rm DP}_{[j]}(\frac{\alpha}{J^t}, \mathbf{H}).
\end{equation}

In Sec.~\ref{subsec:local_inference}, we will explain the sequential inference conducted within each processor that turns the problem into an adaptive component assignment progress. New Gaussian components will be instantiated on-the-fly with knowledge learned from previous observations, guaranteeing an adaptive number of components locally under a globally consistent constraint.
\section{Implementation}
\label{sec:implementation}
\begin{figure*}[htb]
	\begin{center}
		\includegraphics[width=0.95\linewidth]{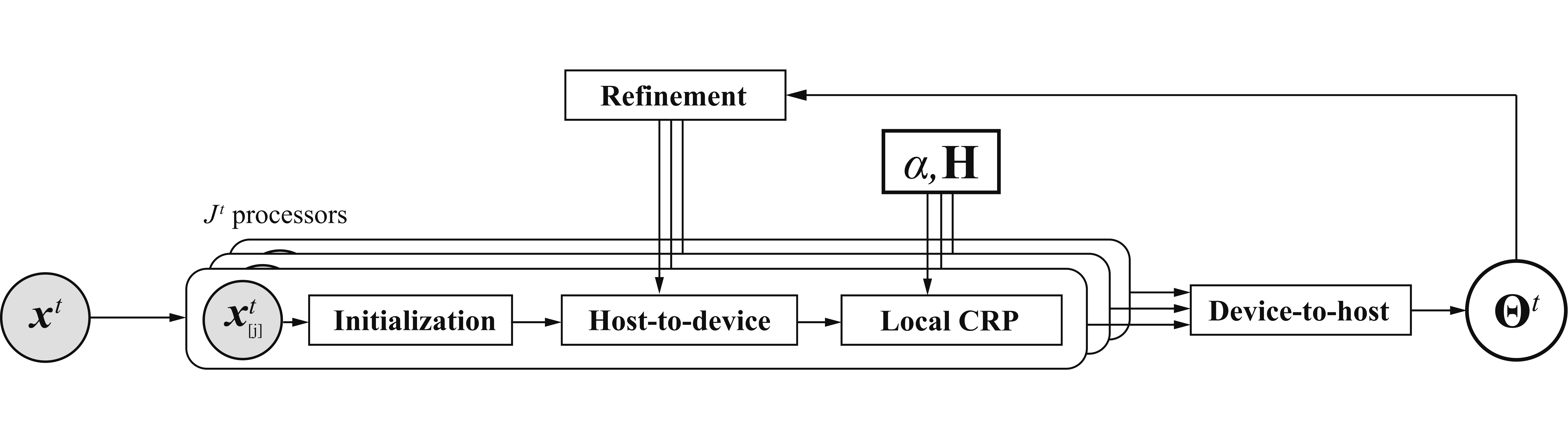}
	\end{center}
	\caption{The proposed online Bayesian learning pipeline. The data are assigned to local processors for parallel updating, where local mixtures are constrained by the global hyper-parameters $\{\alpha, \textbf{H}\}$. The local mixtures are then reweighted and refined as the updated global parameters.}
	\label{fig:inference}
\end{figure*}
Our pipeline is illustrated in Fig.~\ref{fig:inference}. The obtained data are first assigned to different processing units (Sec.~\ref{subsec:hashing}). Afterwards, local DP is inferred in parallel constrained by hyper-priors. Learned parameters are then streamed to host, reweighted and refined as the map measurement $\boldsymbol{\Theta}^t$.

\subsection{Initialization}
\label{subsec:hashing}
We specify the processor indicator $\pi_i^t$ that distributes data mini-batches at each time to $J^t$ processors using spatial hashing algorithm~\cite{VoxelHashing}. Spatial hashing guarantees an $O(1)$ indexing from the coordinate $\mathbf{x}_i^t=(x,y,z)$ to the corresponding processor as:
\begin{equation}
	\label{eq:parallel}
	H(x,y,z)=(x\cdot p_1\oplus y\cdot p_2 \oplus z \cdot p_3)\ \mathit{mod}\ n.
\end{equation}

We follow VoxelHashing~\cite{VoxelHashing} to subdivide the space into voxel blocks, where each block contains $8^3$ voxels. New blocks will be allocated once it falls into the footprint of a new observation. We adopt a lock-based block allocation~\cite{Dong18icra} to avoid thread conflicts. 3D data that are associated with the same mixture component share the same processor indicator, where each processor maintains multiple components $\boldsymbol{\theta}_k^t$ that are corresponding to the same local DP.

\subsection{Local Inference}
\label{subsec:local_inference}
The local inference is conducted in parallel between processors. We here resort to a Chinese restaurant process (CRP) implementation to incrementally update the local DP. By marginalizing over the infinite length partitions for $\boldsymbol{\theta}_k^t$, the parameter updating can be viewed as a procedure of adding and refining mixture components on-the-fly when needed, which resembles the fusion-based map updating in a globally consistent manner. 

Inference with CRP is trivial by first calculating the component assignment $z_i^t \in \mathbf{Z}^t$ and then updating the parameters $\boldsymbol{\Theta}^t$. Component assignment is done parallelly within the associated processor by assigning the point to an existing component k with the probability of:
\begin{equation}
	\label{eq:assignment}
	p(z_i=k|\mathbf{Z}_{-i},\alpha)=\frac{n_{-i,k}}{n-1+\frac{\alpha}{J^t}},
\end{equation}
or instantiating a new component $K^t+1$ with:
\begin{equation}
	\label{eq:new_component}
	p(z_i=K^t+1|\mathbf{Z}_{-i},\alpha)=\frac{\frac{\alpha}{J^t}}{n-1+\frac{\alpha}{J^t}},
\end{equation}
where the subscript $-i$ denotes all indices before $i$th point arrives. $n_{-i,k}$ denotes the number of data that are marginalized out by all mixture components within the processor before $i$th point arrives.

After assigning the data to a specific component, the component parameter can be incrementally updated following \cite{Haines2013pami} as:
\begin{align}
		\label{eq:updating}
		\omega_{i,n_k+1}&=\omega_{i,n_k}+1,\\
		\boldsymbol{\mu}_{i,n_k+1}&=\frac{\omega_{i,n_k}\boldsymbol{\mu}_{i,n_k}+\mathbf{x}_i}{\omega_{i,n_k}+1},\\
		\boldsymbol{\Sigma}_{i,n_k+1}&=\boldsymbol{\Sigma}_{i,n_k}+\frac{\omega_{i,n_k}}{\omega_{i,n_k}+1}(\mathbf{x}_i-\boldsymbol{\mu}_{i,n_k})^2.
\end{align}
\subsection{Map Refinement}
Though the hyper-priors for CRP enforce component instantiation when needed, the inference within a processor is conducted sequentially. Hence, a large amount of newly instantiated components may make the system computationally intractable even with a GPU acceleration. Practically, we perform truncation and pruning to maintain a clean and compact parameter space. Truncation is implemented by setting an upper bound of the mixture number for each processor as $T$. By enforcing $p(z_i=k)=0$ for $k\geq T$, memory pre-allocation and fast indexing is guaranteed within each processor.

\begin{figure}[!b]
	\begin{center}
		\subfloat[Samples from original map]{
			\includegraphics[width=0.45\linewidth]{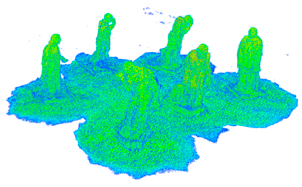}
			\label{subfig:MRFmap}
		}
		\subfloat[Samples from pruned map]{
			\includegraphics[width=0.45\linewidth]{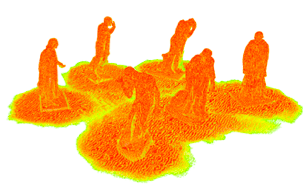}
			\label{subfig:kdndt}
		}
	\end{center}
	\caption{Map refinement from noisy RGB-D data stream. }
	\label{fig:noisy}
\end{figure}

On the other hand, it is still possible that some of the components are redundant. We follow the Sequential Variational Approximation (SVA)~\cite{Lin2013nips} to explicitly maintain an accumulated weight for each component and adopt a thresholding pruning when necessary. The weight takes both point-Gaussian distance and data fidelity into consideration. An example of map updating from noisy RGB-D data is illustrated in Fig.~\ref{fig:noisy}. Our strict component instantiation strategy guarantees a clean and compact mixture of Gaussians. 

To measure the data fidelity, we here view acquired 3D data as Gaussian-distributed noisy observations from samples of the global distribution as $\hat{\mathbf{x}_i^t} \sim \mathbf{\mathcal{N}}(\mathbf{x}_i^t, \mathbf{\Sigma}_{\mathbf{x}_i^t})$, which can be optionally replaced by a specific model for a particular sensor input such as \cite{Nguyen2012Kinect} and \cite{Huber2018icra}. Following \cite{Cao2018tog, Dhawale2020rss}, the covariance is represented as:
\begin{equation}
	\mathbf{\Sigma}_{\mathbf{x}} = \mathbf{J}_{\mathbf{x}}\cdot{\rm diag}(\sigma_u^2, \sigma_v^2, \sigma_z^2)\cdot \mathbf{J}_{\mathbf{x}}^T,
\end{equation}
where $\sigma_u^2$and $\sigma_v^2$ are pixel-positional variance set to be half-pixel size $0.5^2$. $\sigma_z^2$ is the depth variance obtained by \cite{Nguyen2012Kinect}. $\mathbf{J}_{\mathbf{x}}$ is the Jacobian matrix as:
\begin{equation}
	\mathbf{J}_{\mathbf{x}}={
		\left[ \begin{array}{ccc}
			f_x^{-1} & 0 & (u-c_x)f_x^{-1}\\
			0 & f_y^{-1} & (v-c_y)f_y^{-1}\\
			0 & 0 & 1
		\end{array} 
		\right ]}.
\end{equation}

\section{Experiments}
In this section, we first present our experimental setup and evaluation protocols. Afterwards, we compare our representation against other related representations with state-of-the-art performances in terms of accuracy and efficiency. Qualitative results of the proposed representation on different datasets are also presented.
\subsection{Experimental Setup}
The experiments were conducted on a desktop PC with an Intel Core i7-6700 (8 cores @ 4 GHz), 16GB of RAM, and a single NVIDIA GeForce RTX 2080Ti with 11GB of memory.

\noindent\textbf{State-of-the-art methods.} Due to our spatial hashing implementation, we compare our method against state-of-the-art voxel-based probabilistic representations that can be updated in real-time. Default parameters are taken for each to perform confidence reasoning and outlier rejection.
\begin{itemize}[nosep,leftmargin=\parindent,itemindent=0pt]
	\item MRFMap\footnote{\url{https://github.com/mrfmap/mrfmap}}~\cite{Shankar2020rss} maintains a forward ray sensor model via a Markov random field. The experiments are conducted on selected keyframes due to the increasing computational cost along with the graph size.
	\item  KD-NDT\footnote{\url{https://github.com/cogsys-tuebingen/cslibs\_ndt}}~\cite{Schulz2018iros} maintains local Gaussian distribution within overlapped grid cell indexed by multiple kd-trees to mitigate the discretization error induced by voxelization. The CPU implementation without parallelization makes the method computationally intractable. The experiments are conducted on selected keyframes with downsampled depth data.
	\item  PSDF\footnote{\url{https://github.com/theNded/MeshHashing}}~\cite{Dong2018eccv} maintains a joint distribution of SDF value and its inlier probability and outperforms traditional TSDF-fusion methods with noise and outlier handling.
	\item  RoutedFusion\footnote{\url{https://github.com/weders/RoutedFusion}}~\cite{Weder2020cvpr} trains a 2D depth routing network and a 3D depth fusion network to handle anisotropically distributed data fusion. We use the pre-trained model for evaluation.
\end{itemize}

\noindent\textbf{Metrics.} We evaluate our representation quality by computing the mean and the standard deviation (std.) of the cloud-mesh distance using \emph{CloudCompare}\footnote{\url{http://www.danielgm.net/cc/}} software. Since output formats vary between different representations, we randomly sample 150,000 points from each representation for quantitative evaluation. As KD-NDT only outputs means of Gaussians, we directly take downsampled mean values as sampled points. For our representation, samples are generated using importance sampling~\cite{Eckart2016cvpr}, which approximates the global distribution and is noisier compared to the mean value.

\noindent\textbf{Datasets.} We mainly evaluate quantitatively on the synthetic ICL-NUIM livingroom dataset~\cite{ICL}. Additional evaluations are performed on TUM RGB-D Dataset (TUM)~\cite{TUM} and 3D Scene Data (Zhou)~\cite{Zhou} with real scans.

\begin{figure}[!tbh]
	\centering
	\subfloat[Augmented ICL]{
		\includegraphics[width=0.30\linewidth]{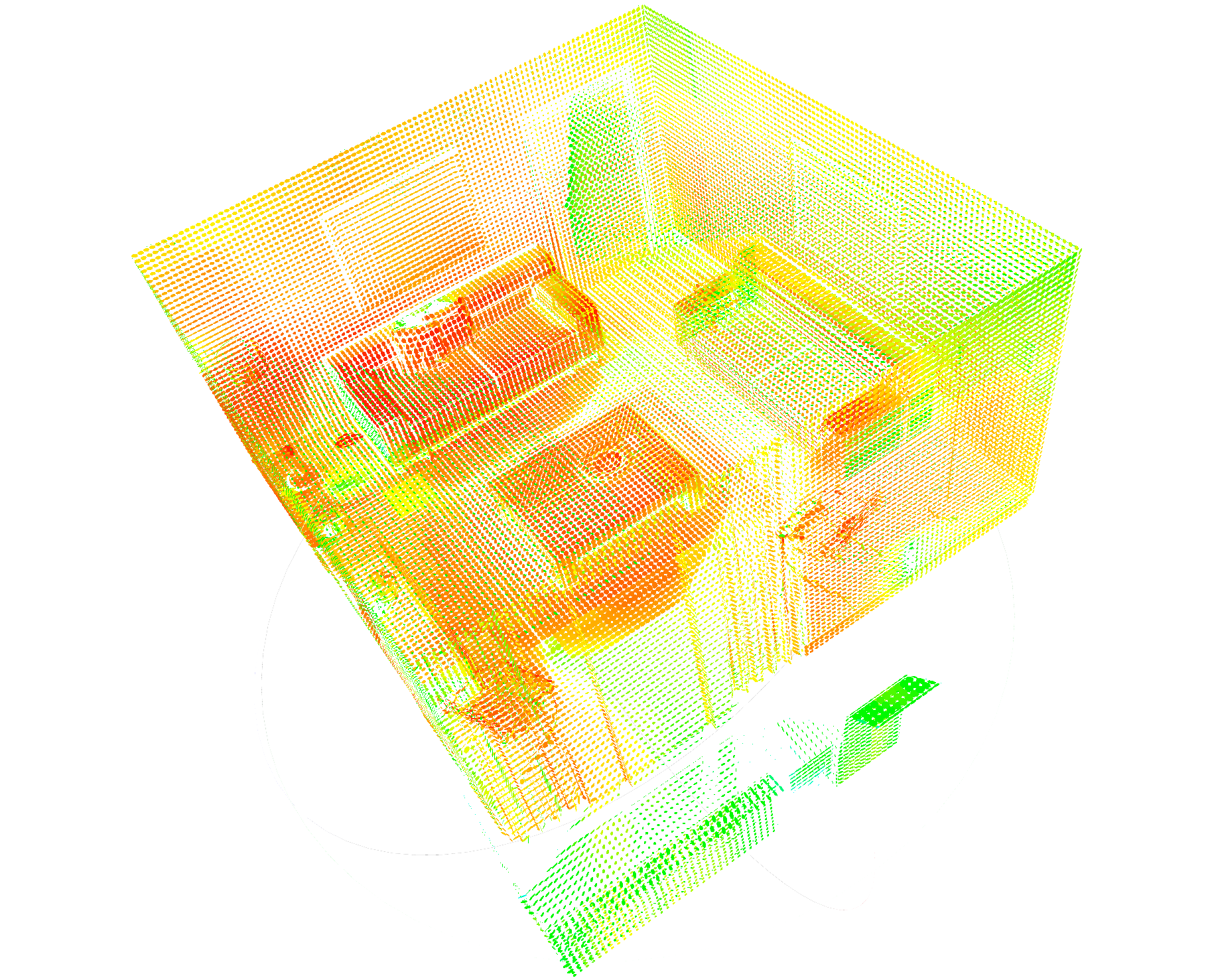}
		\label{subfig:kdndt}
	}
	\subfloat[Copyroom]{
		\includegraphics[width=0.30\linewidth]{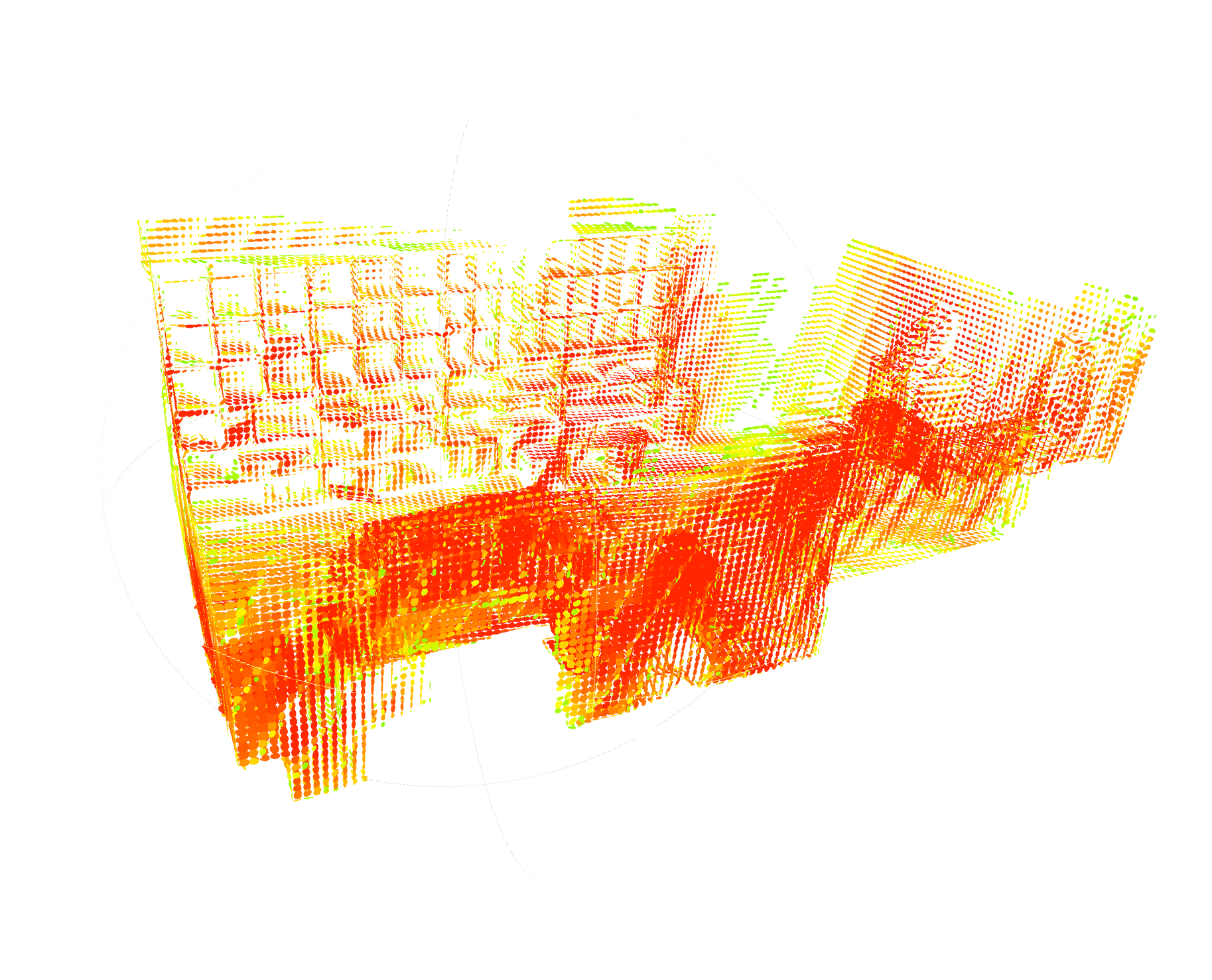}
		\label{subfig:psdf}
	}
	\subfloat[Lounge]{
		\includegraphics[width=0.30\linewidth]{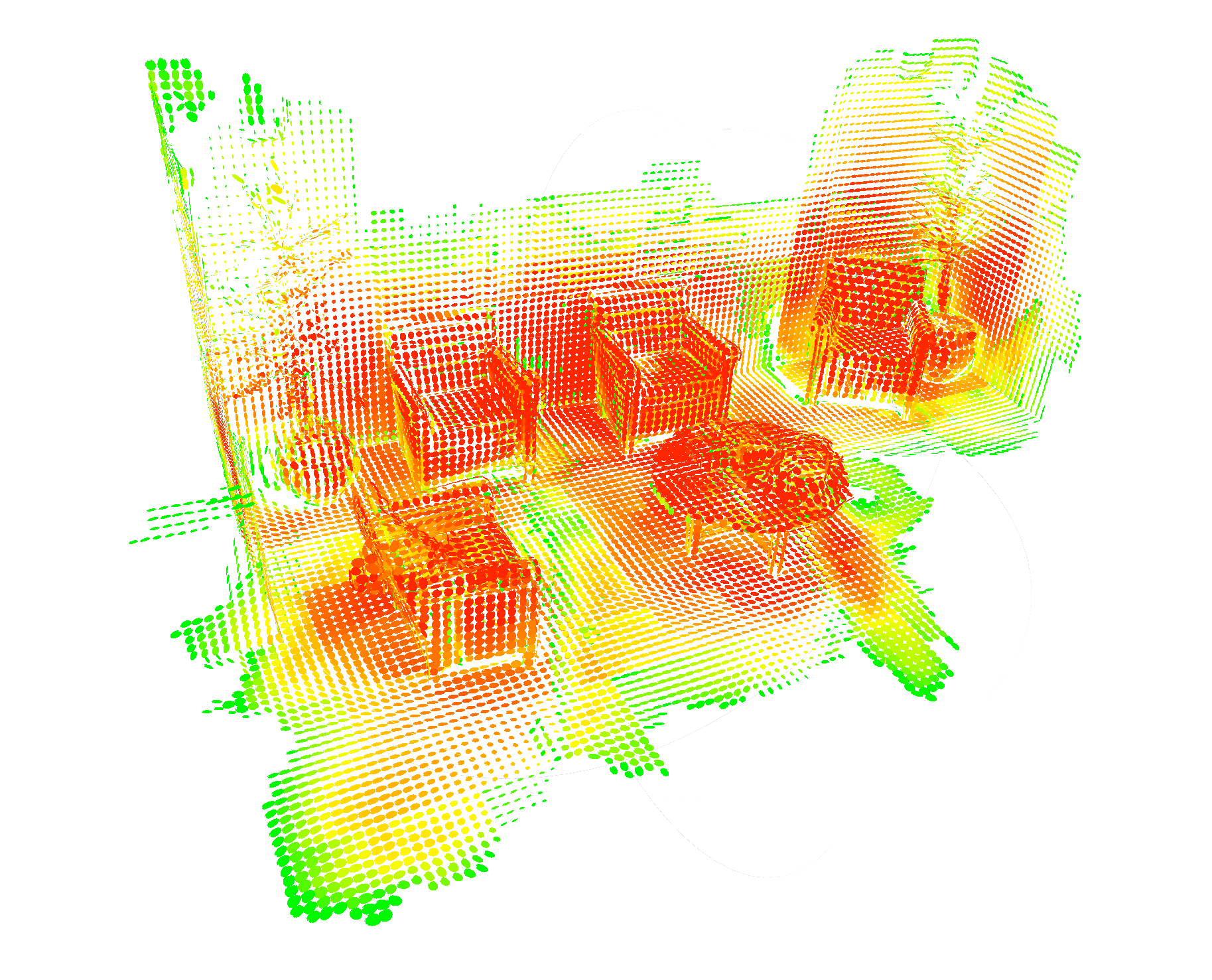}
		\label{subfig:routedfusion}
	}\\
    \subfloat[fr1\_xyz]{
    	\includegraphics[width=0.30\linewidth]{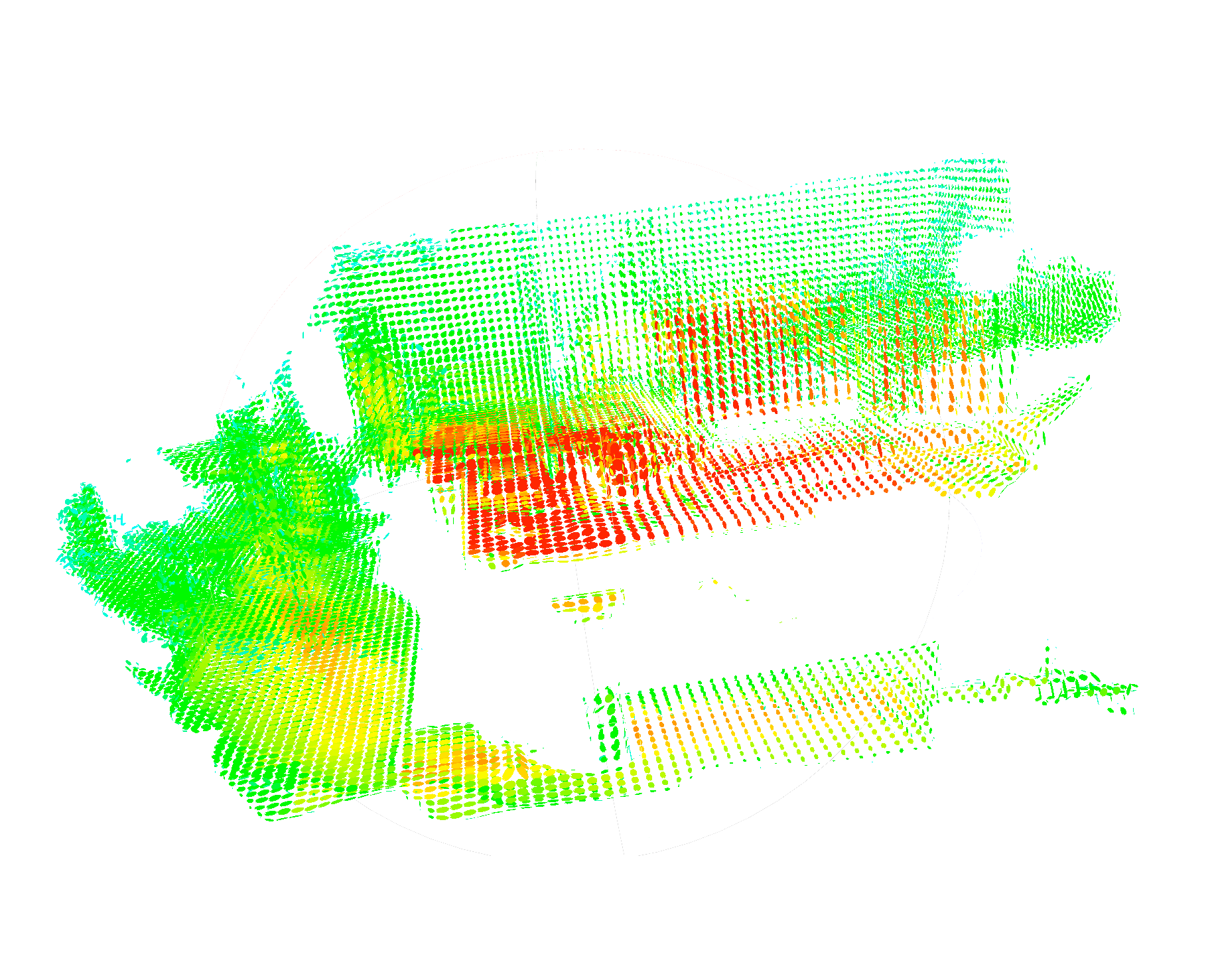}
    	\label{subfig:psdf}
    }
    \subfloat[fr2\_xyz]{
    	\includegraphics[width=0.30\linewidth]{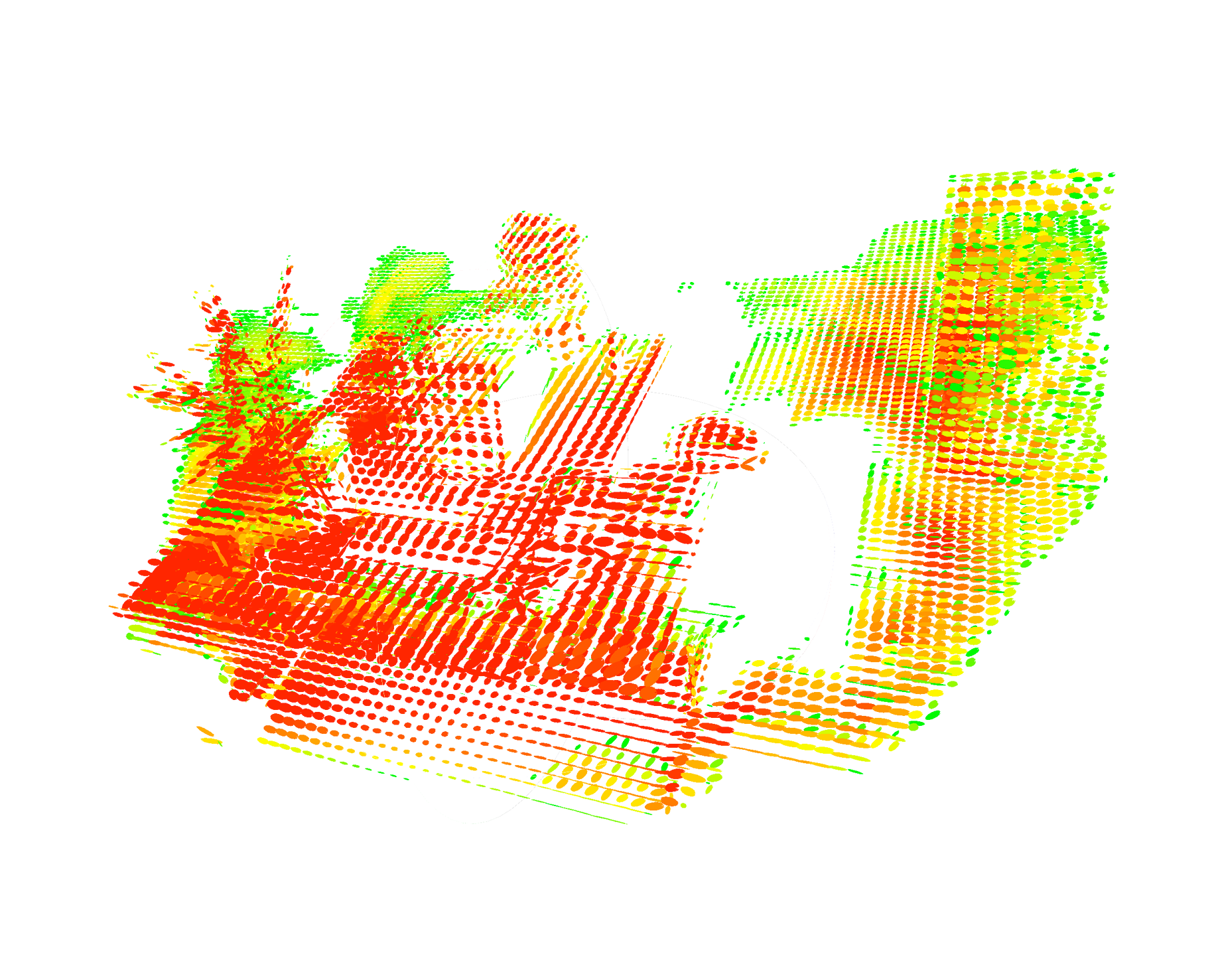}
    	\label{subfig:routedfusion}
    }
    	\subfloat[fr3\_long\_office]{
    	\includegraphics[width=0.30\linewidth]{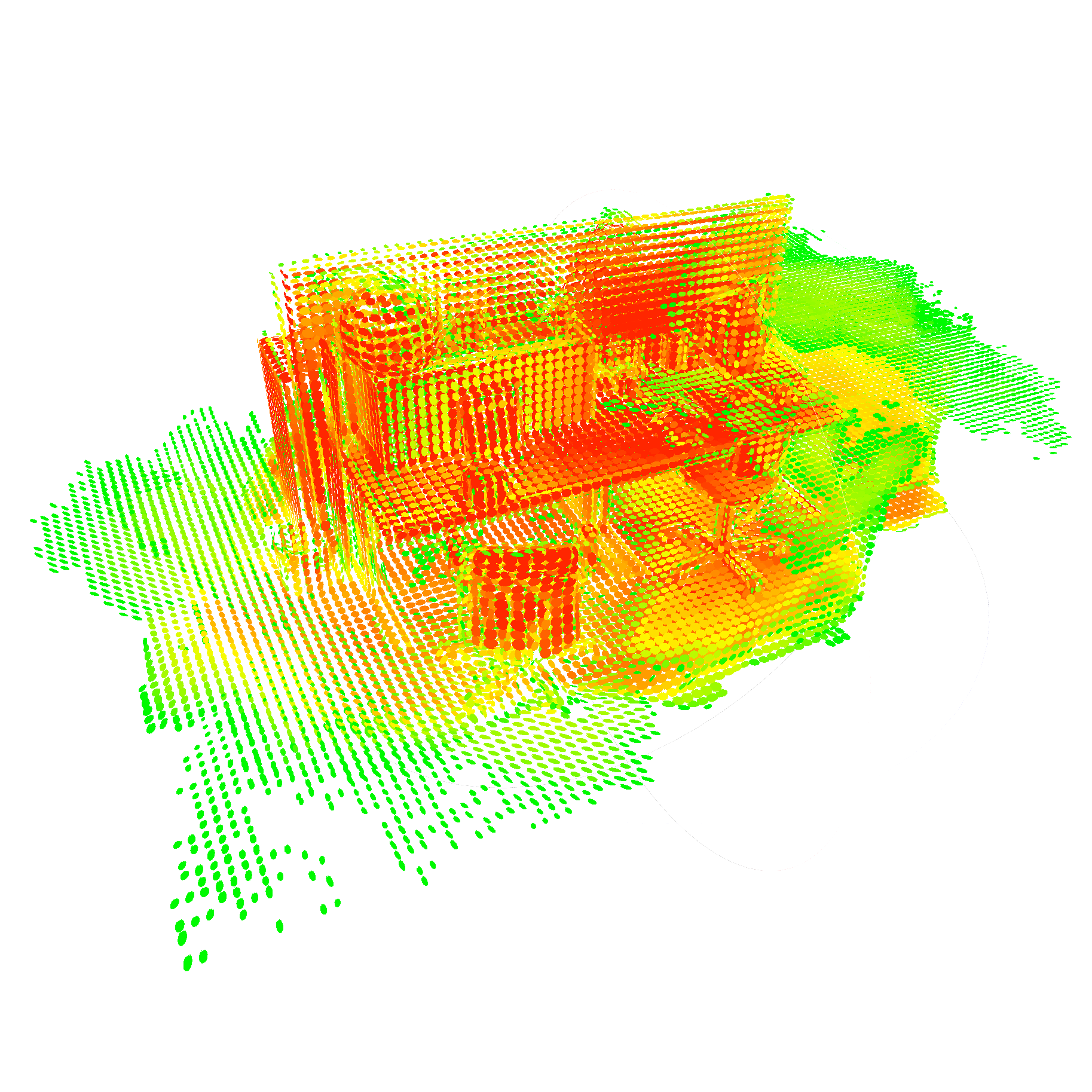}
    	\label{subfig:MRFmap}
    }
	\caption{Visualization of the proposed scene representation on various datasets. Color denotes the local accumulated confidence (red→blue: high→low confidence).}
	\label{fig:vis}
\end{figure}
\subsection{Representation Quality}
\begin{figure*}[!h]
	\centering
	\subfloat[MRFMap]{
		\includegraphics[width=0.19\linewidth]{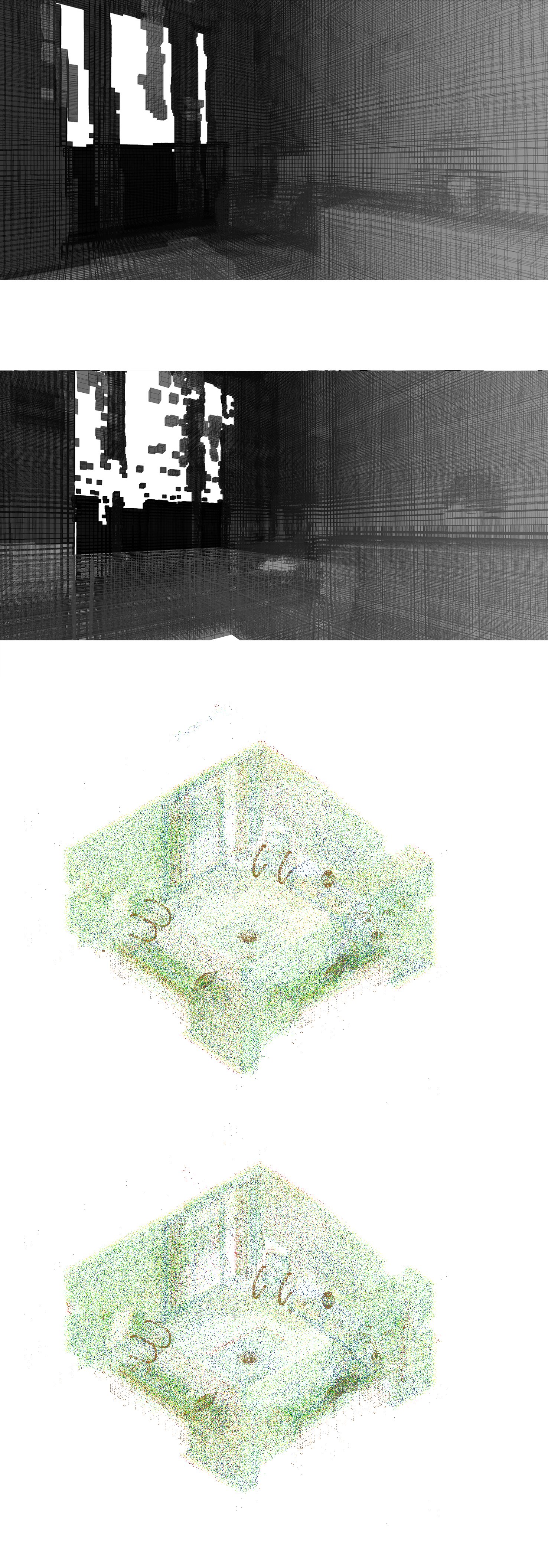}
		\label{subfig:MRFmap}
	}
	\subfloat[KD-NDT]{
		\includegraphics[width=0.19\linewidth]{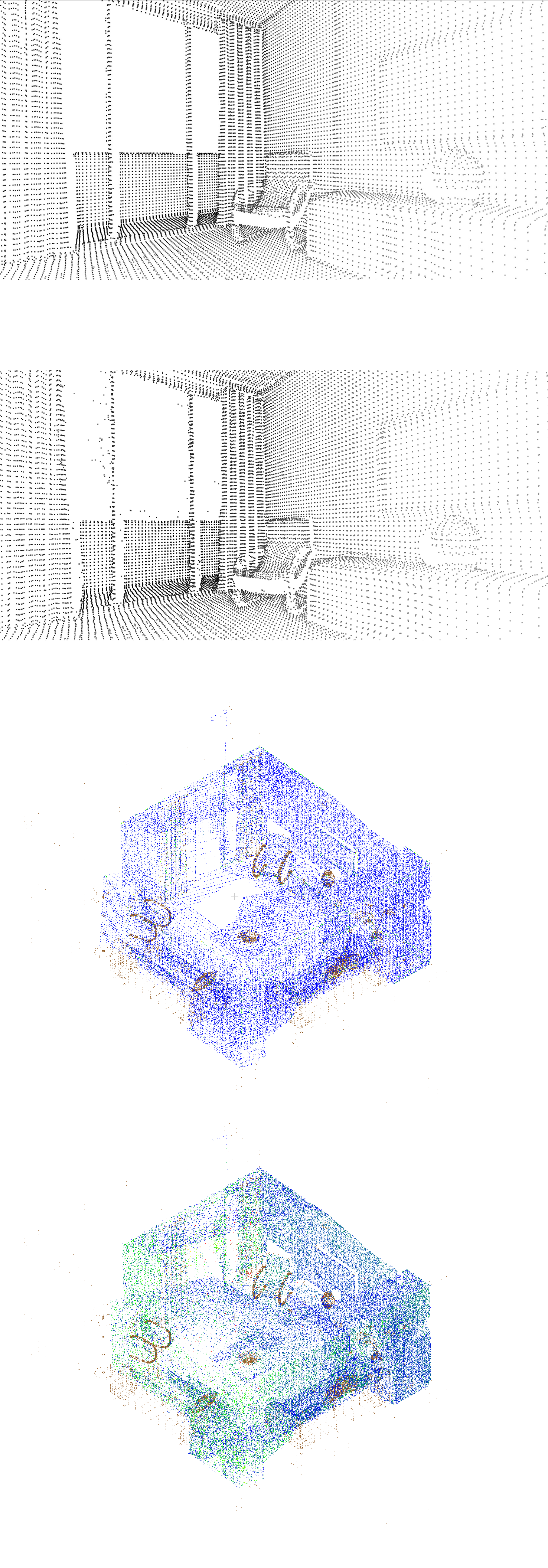}
		\label{subfig:kdndt}
	}
	\subfloat[PSDF]{
		\includegraphics[width=0.19\linewidth]{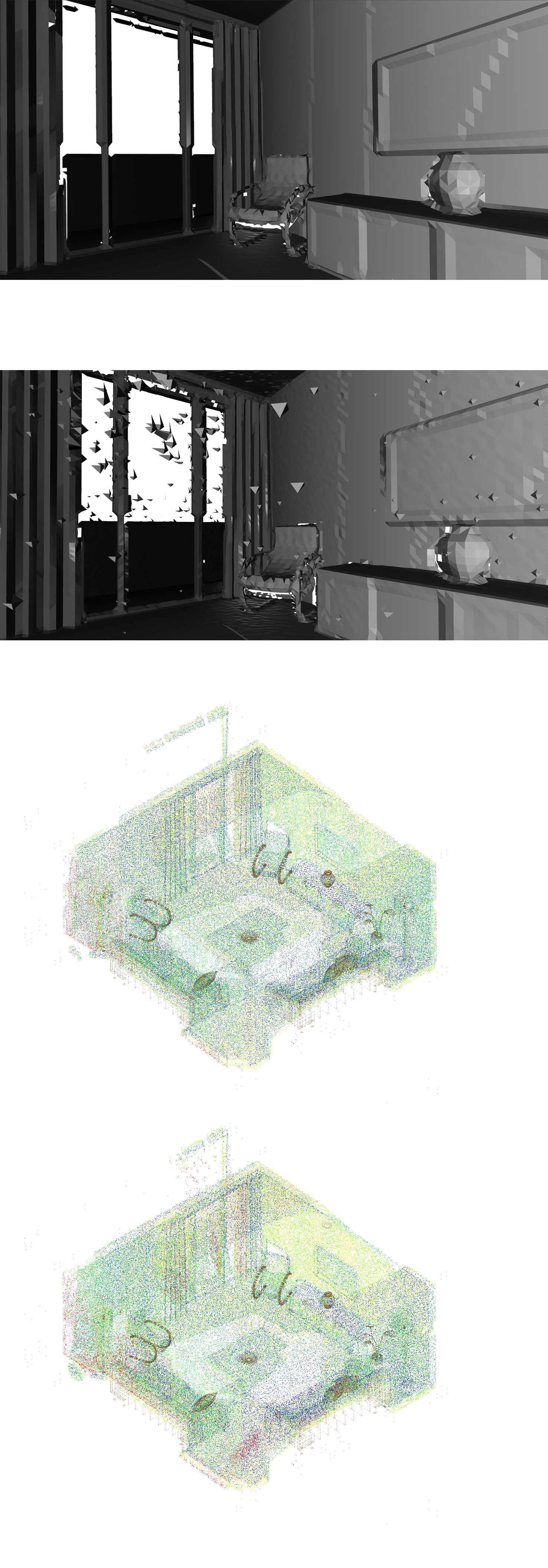}
		\label{subfig:psdf}
	}
	\subfloat[RoutedFusion]{
		\includegraphics[width=0.19\linewidth]{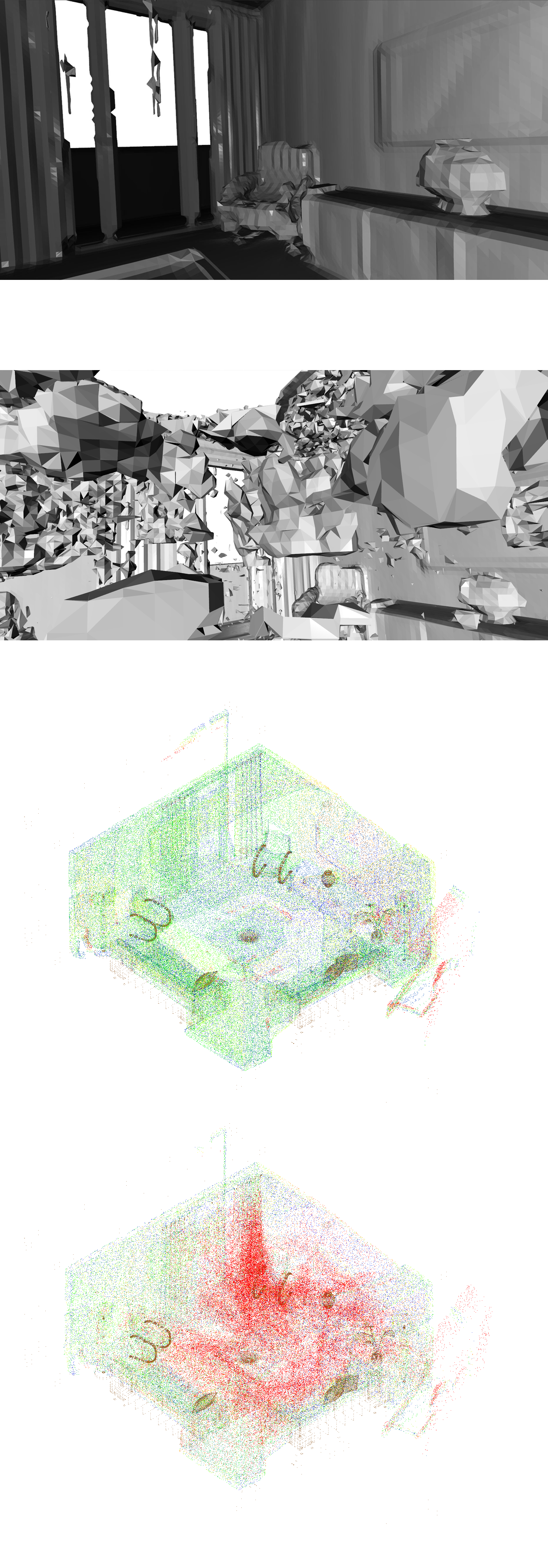}
		\label{subfig:routedfusion}
	}
	\subfloat[Ours]{
		\includegraphics[width=0.19\linewidth]{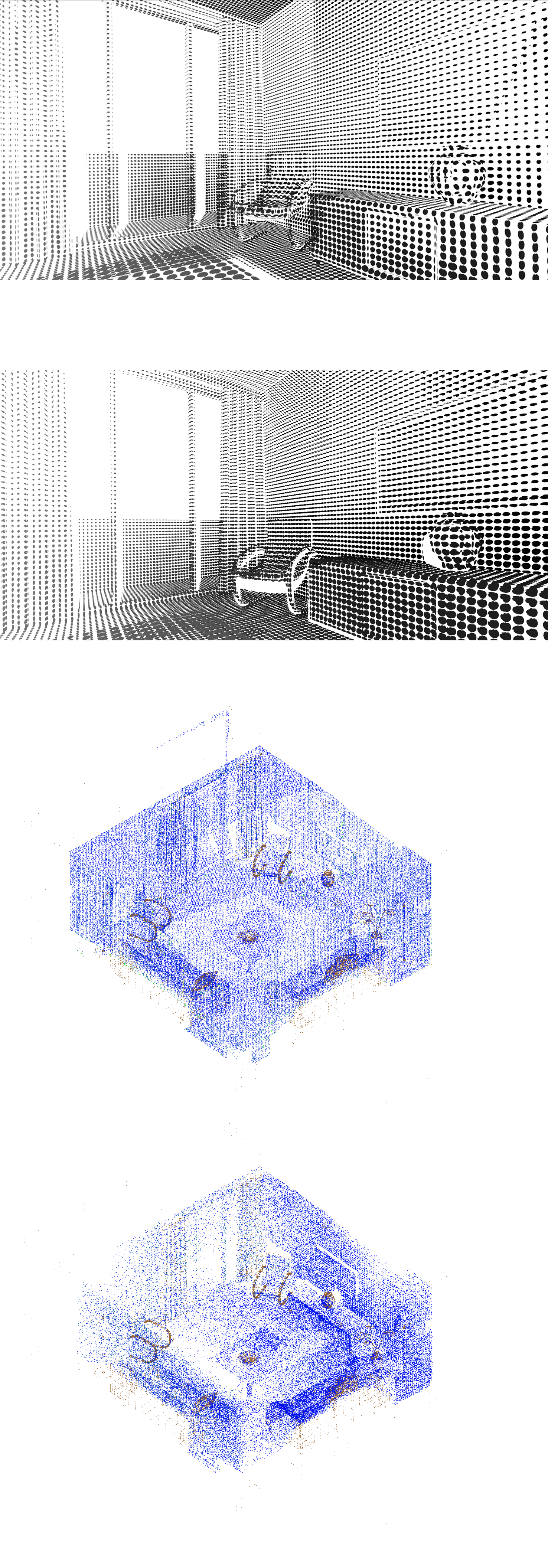}
		\label{subfig:ours}
	}
	\caption{Qualitative comparisons of the representation quality. First two rows: visualization of each scene representation on kt1 clean/noisy sequences. The voxel resolution is set as 4 cm. Last two rows: Error heatmap on kt2 clean/noisy sequences. Color denotes the point-to-mesh distance (blue-red: 0-10cm). The voxel resolution is set as 2cm.}
	\label{fig:accuracy}
\end{figure*}

\setlength{\tabcolsep}{4pt}
\begin{table*}[htb]
	\renewcommand\arraystretch{1.1}
	
	\begin{center}
		\caption{Cloud-to-mesh statistics (cm) on the ICL-NUIM dataset. The voxel resolution is set as 5cm.}
		\label{tab:accuracy}
		 \setlength{\tabcolsep}{3mm}
		\begin{tabular}{|l|c|c|c|c|c|c|c|c|}
			\hline
			\multirow{2}{*}{Method} &
			\multicolumn{2}{|c|}{kt1 (\emph{clean})} &
			\multicolumn{2}{|c|}{kt1 (\emph{noisy})} &
			\multicolumn{2}{|c|}{kt2 (\emph{clean})} &
			\multicolumn{2}{|c|}{kt2 (\emph{noisy})}\\
			\cline{2-3} \cline{4-5} \cline{6-7} \cline{8-9}
			& mean & std. & mean & std. & mean & std. &mean & std.\\
			\hline\hline
			MRFMap~\cite{Hornung2013ar} & 5.3759& 3.3863& 5.3028& 4.9927&5.3979 &3.3210 &5.2735 &4.4339 \\
			KD-NDT~\cite{Schulz2018iros} & 0.2675& 0.4948& 1.0688& 1.2475& 0.2492& 0.4603& 1.4877& 1.8574\\
			PSDF~\cite{Dong2018eccv} & 3.5104& 3.4741& 5.5026& 9.9667&4.2282 & 3.6356& 5.2325&6.1295 \\
			RoutedFusion~\cite{Weder2020cvpr} & 6.5169& 3.3753& 20.3565 & 19.1897 & 5.0746& 3.1432& 23.7922& 27.0850 \\
			Ours & \textbf{0.0752} & \textbf{0.1321}& \textbf{0.8709}&\textbf{1.0549} &\textbf{0.0659} & \textbf{0.1195}& \textbf{1.0078} & \textbf{0.8658} \\
			\hline
		\end{tabular}
	\end{center}
\end{table*}
\setlength{\tabcolsep}{1.4pt}
We conduct qualitative and quantitative evaluation to measure how well the proposed representation can describe the generative process of a scene. The visualization of the proposed representation is illustrated in Fig.~\ref{fig:vis}. We compare against other representations on both clean and noisy sequences of the ICL-NUIM dataset to further demonstrate the noise-handling ability of each representation. As shown in Tab.~\ref{tab:accuracy} and Fig.~\ref{fig:accuracy}, our representation achieves a much lower error compared to other baselines. We provide various voxel resolution configurations and obtain consistent findings:

1) Descriptiveness: It is noteworthy that at a low voxel resolution, Gaussian-based representations such as KD-NDT and ours are more capable of modeling clean and thin surfaces. As we maintain an adaptive number of Gaussians within each voxel grid, the representation is more descriptive compared to the NDT-based representations. Even though our parameter space is sparse and discrete, the representation itself is a continuous probability field. Hence, the map serves as a generative model where we can sample arbitrary number of points. Besides, the sampled density reflects the local geometric confidence (Fig.~\ref{subfig:ours}).

2) Noise-handling ability: It can also be noted that our representation achieves the lowest error and deviation on noisy sequences compared to other competitive representations. The systematically established probabilistic formulation along with the truncation and pruning strategies
guarantee a promising accuracy. We clarify that we do not train networks provided by RoutedFusion as we target an online learning fashion. Quantitative evaluation with re-trained networks for RoutedFusion is demonstrated in the supplementary materials.

\subsection{Representation Efficiency}
\begin{figure*}[!bht]
	\centering
	\subfloat[Accuracy vs. runtime]{
	    \includegraphics[width=0.43\linewidth]{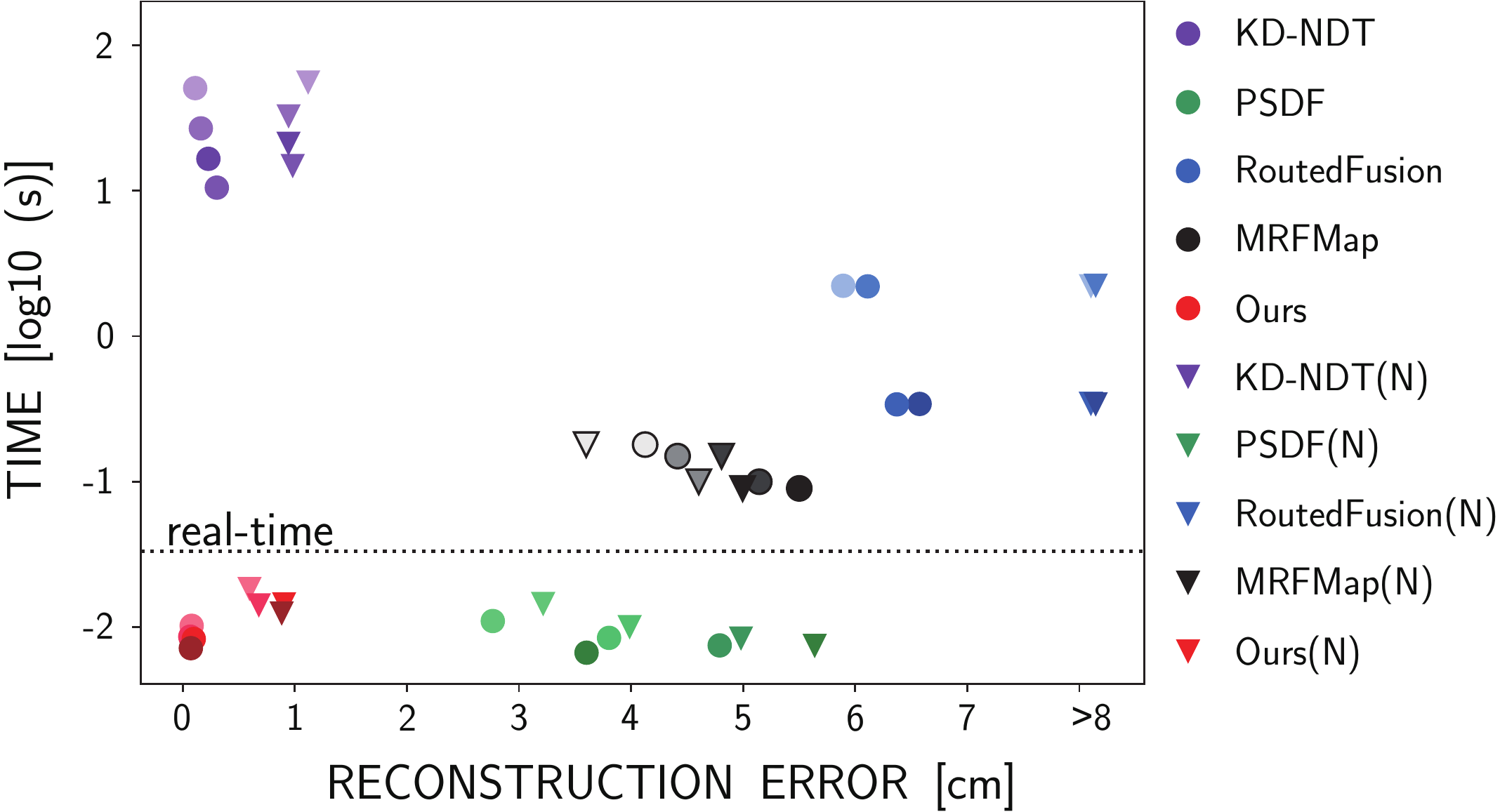}
		\label{subfig:runtime}
	}
    \hspace{10mm}
	\subfloat[Accuracy vs. memory]{
		\includegraphics[width=0.43\linewidth]{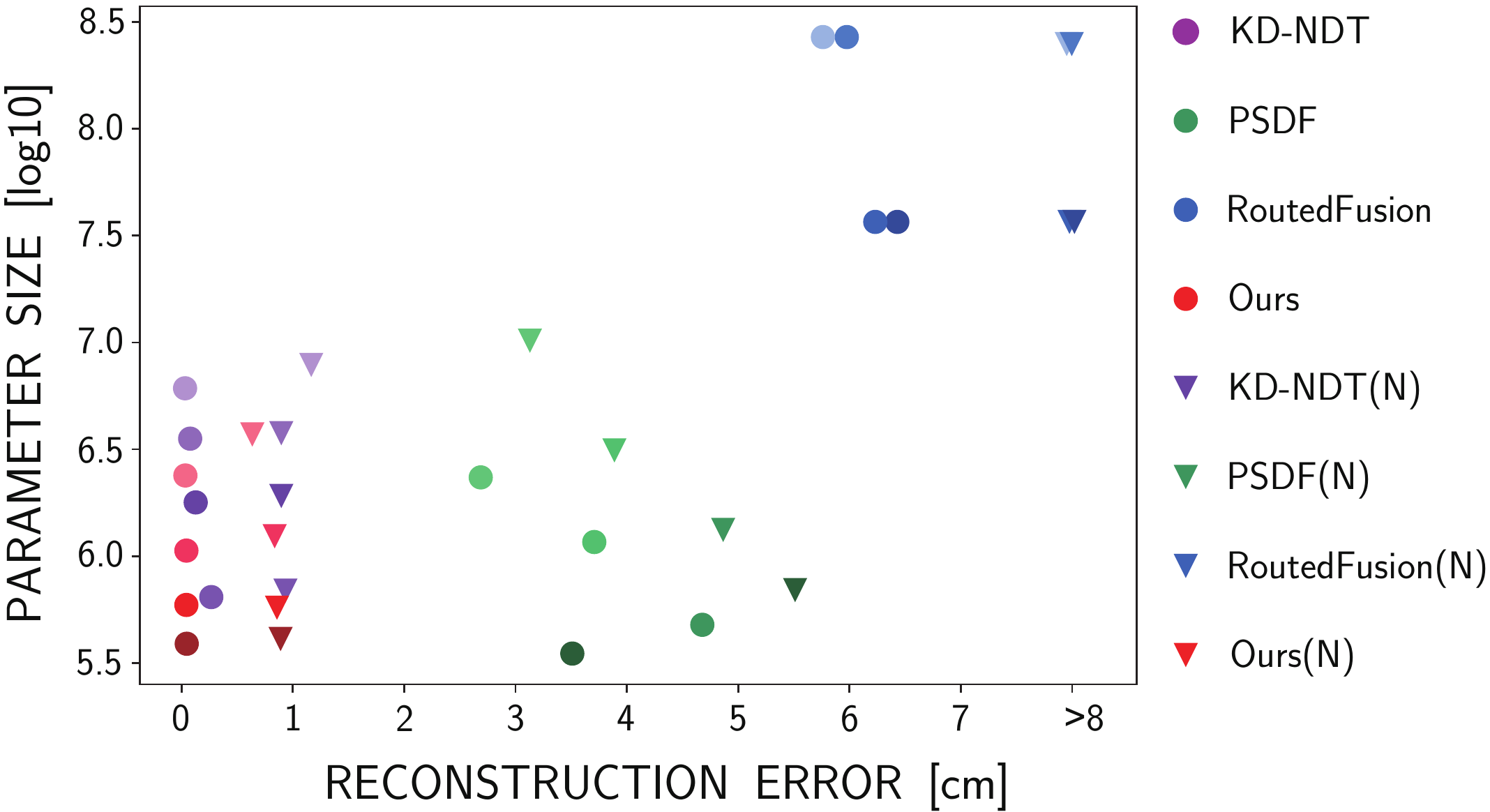}
		\label{subfig:memory}
	}
	\caption{Representation efficiency on ICL-NUIM kt1 sequence. For each method, we evaluate the trade-offs between accuracy-time and accuracy-parameter number on clean and noisy [N] datasets. The voxel resolution is set to be $2$cm-$5$cm, where the lighter color denotes a higher resolution.}
	\label{fig:parameter-accuracy}
\end{figure*}
\begin{figure*}[!h]
	\begin{center}
		\includegraphics[width=0.8\linewidth]{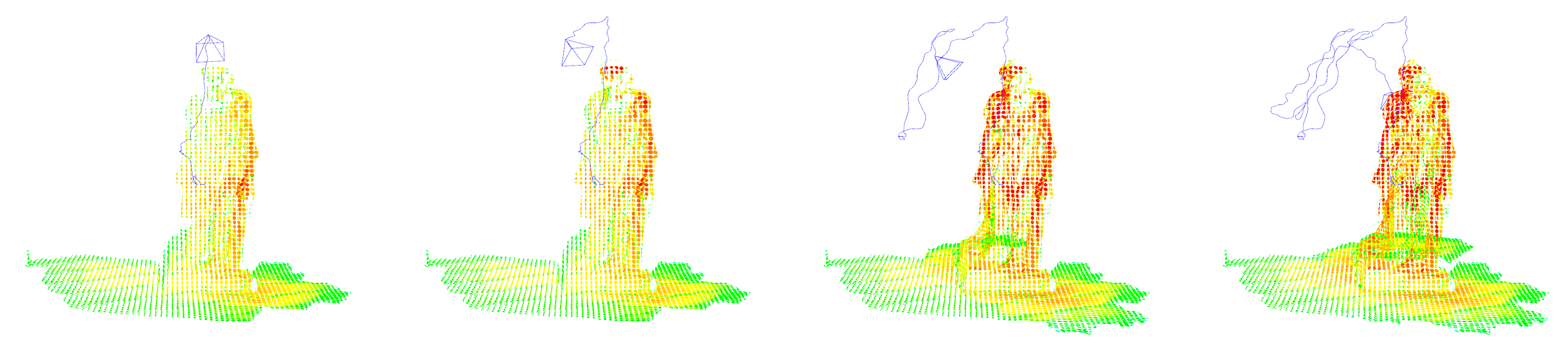}
	\end{center}
	\caption{Online learning of the Bayesian nonparametric model. The confidence and the completeness are gradually increasing by gaining knowledge from streaming data.}
	\label{fig:online_learning}
\end{figure*}
We measure the representation efficiency in terms of accuracy \vs runtime/parameter number at different voxel resolution (2cm-5cm) configurations. All baselines except KD-NDT are implemented in parallel on a single GPU. As illustrated in Fig.~\ref{subfig:runtime}, the proposed method yields a good trade-off between accuracy and computational efficiency. Our spatial hashing scheme is similar to PSDF~\cite{Dong2018eccv}. Though our local sequential inference leads to additional computational cost compared to SDF updating, we achieve efficient inference at a high voxel resolution. It can be explained twofold: Firstly, new components can hardly be instantiated as high voxel resolution leads to a large $J^t$ and in turn a small $\frac{\alpha}{J^t}$. Secondly, the size of minibatch data within each processor is small, thus leading to lower complexity of local sequential inference. It should also be noted that the computational and memory efficiency of RoutedFusion is up to the defined volume size. For livingroom dataset at the size of about $6$m*$3$m*$9$m, the voxel grid is allocated to be $256^3$ for $4$cm and $5$cm resolution and $512^3$ for $2$cm and $3$cm. 

Furthermore, Fig.~\ref{subfig:memory} depicts the trade-off between memory consumption and accuracy. It can be noticed that we achieve high accuracy at a relatively low memory consumption. The parameter size is calculated by multiplying the allocated voxel number by the parameter number within each voxel. It should be noted that we do not provide the parameter size of MRFMap. Keyframe sensor data are required to be stored to construct the MRF graph. Hence, the memory consumption will increase monotonically when adding more keyframes. Meanwhile, the implemented overlapping grids for KD-NDT lead to an effective resolution of half the resolution of the voxel grid, thus bringing more Gaussian components compared against ours.

\section{Conclusion and Future Work}
In this paper, we introduce a Bayesian nonparametric mixture model as the scene representation, depicting a continuous probability density function. Map updating given streaming data is cast as an online Bayesian learning problem,  as illustrated in Fig.~\ref{fig:online_learning}. A gradual transition from geometry prior to posterior is conducted through parallel and incremental inference in real-time. Experimental results demonstrate that the proposed method achieves state-of-the-art accuracy and efficiency.

We believe that the proposed approach establishes a systematical framework based on probabilistic formulation, revealing potentials for multiple extensions. One interesting direction lies in online learning of scene geometry with neural networks. The proposed approach models the transition from geometry prior to posterior and opens the gate to enforce knowledge transfer~\cite{Lee2019iclr} from pre-trained features. Recent advances in learning local geometry primitives may obtain a more expressive prior distribution compared to the assumed Gaussian or other distributions. Another direction lies in the graphical applications derived from the proposed representation. As stated in SurfelMeshing~\cite{SurfelMeshing}, online meshing directly from point-wise data is susceptible to noise. The parameter space of the proposed representation can be directly utilized as a probabilistic surface element that is robust to different sensor noise. The generative property also allows the generation of different scene representations from the probability field. We believe that the proposed representation will store and provide more informative cues for diverse kinds of applications.

\section{Acknowledgement}
We would like to thank Wei Dong for feedback on the manuscript, Kangjie Zhou and Jiarui Liang for their help with the cloud-mesh distance calculation, and anonymous reviewers for their helpful feedback. This work is supported by the National Key Research and Development Program of China
(2017YFB1002601) and National Natural Science Foundation of China (61632003, 61771026).
{\small

\bibliographystyle{ieee_fullname}
\bibliography{egbib}
}

\end{document}